\documentclass[letterpaper, 10 pt, conference]{ieeeconf}
\usepackage[utf8]{inputenc}

\IEEEoverridecommandlockouts                              
\usepackage{graphicx}  				

\usepackage[croatian]{babel}  
\usepackage[utf8]{inputenc}  	
\usepackage[T1]{fontenc}
\usepackage{ae,aecompl}     	

\usepackage{microtype}				

\usepackage{subfig}

\usepackage{amsmath}

\usepackage{tabularx}
\usepackage{booktabs}
\usepackage{enumerate}				
\usepackage{algorithm2e}			
\usepackage{todonotes}				
\usepackage{dirtree}					
\usepackage{hyperref}					

\usepackage{lmodern}
\usepackage{nccmath}
\usepackage{scalerel,stackengine}
\usepackage{comment}
\usepackage{cite}

\usepackage{amssymb}
\usepackage{makecell}

\usepackage{calrsfs}
\DeclareMathAlphabet{\pazocal}{OMS}{zplm}{m}{n}

\graphicspath{{./figures/}}
\usepackage{float}

\newcommand{\norm}[1]{\left\lVert #1 \right\rVert}

\title{\LARGE \bf
	Constrained Prioritized 3T2R Task Control for Robotic Agricultural Spraying  
}
\author{Ivo Vatavuk, Zdenko Kovačić
	\thanks{\hrule}
	\thanks{Ivo Vatavuk, MSc, is a PhD student at the Faculty of Electrical Engineering and Computing,
	University of Zagreb, Unska 3, 10000 Zagreb, Croatia: \tt(ivo.vatavuk@fer.hr)
	}
	\thanks{
	Zdenko Kovačić, PhD, is a full professor at the Faculty of Electrical Engineering and Computing,
	University of Zagreb, Unska 3, 10000 Zagreb, Croatia \tt(zdenko.kovacic@fer.hr)
	}}%

\makeatletter
\newcommand{\removelatexerror}{\let\@latex@error\@gobble}

\makeatother

\begin{document}
	
	\maketitle
	\thispagestyle{empty}
	\pagestyle{empty}
	
	\begin{abstract}
		In this paper, we present a solution for robot arm-controlled agricultural spraying, handling the spraying task as a constrained prioritized 3T2R task. 3T2R tasks in robot manipulation consist of three translational and two rotational degrees of freedom, and are frequently used when the end-effector is axis-symmetric. The solution presented in this paper introduces a prioritization between the translational and rotational degrees of freedom of the 3T2R task, and we discuss the utility of this kind of approach for both velocity and positional inverse kinematics, which relate to continuous and selective agricultural spraying applications respectively.       
	\end{abstract}
	\section{Introduction}

Agricultural robotics is a rapidly advancing research field that focuses on developing and deploying robotic technology for various agricultural tasks. The goal is to enhance the efficiency and sustainability of different agricultural procedures and address labor shortages. Research presented in this paper is a part of the project HEKTOR \cite{hektor, Goricanec2021}, which aims to introduce heterogeneous robotic systems to the agricultural areas of viticulture and mariculture. A mobile manipulator is envisioned to autonomously perform various viticultural tasks, including monitoring, spraying and suckering.  

\begin{figure}[!ht]
\centering
\includegraphics[width=0.65\columnwidth]{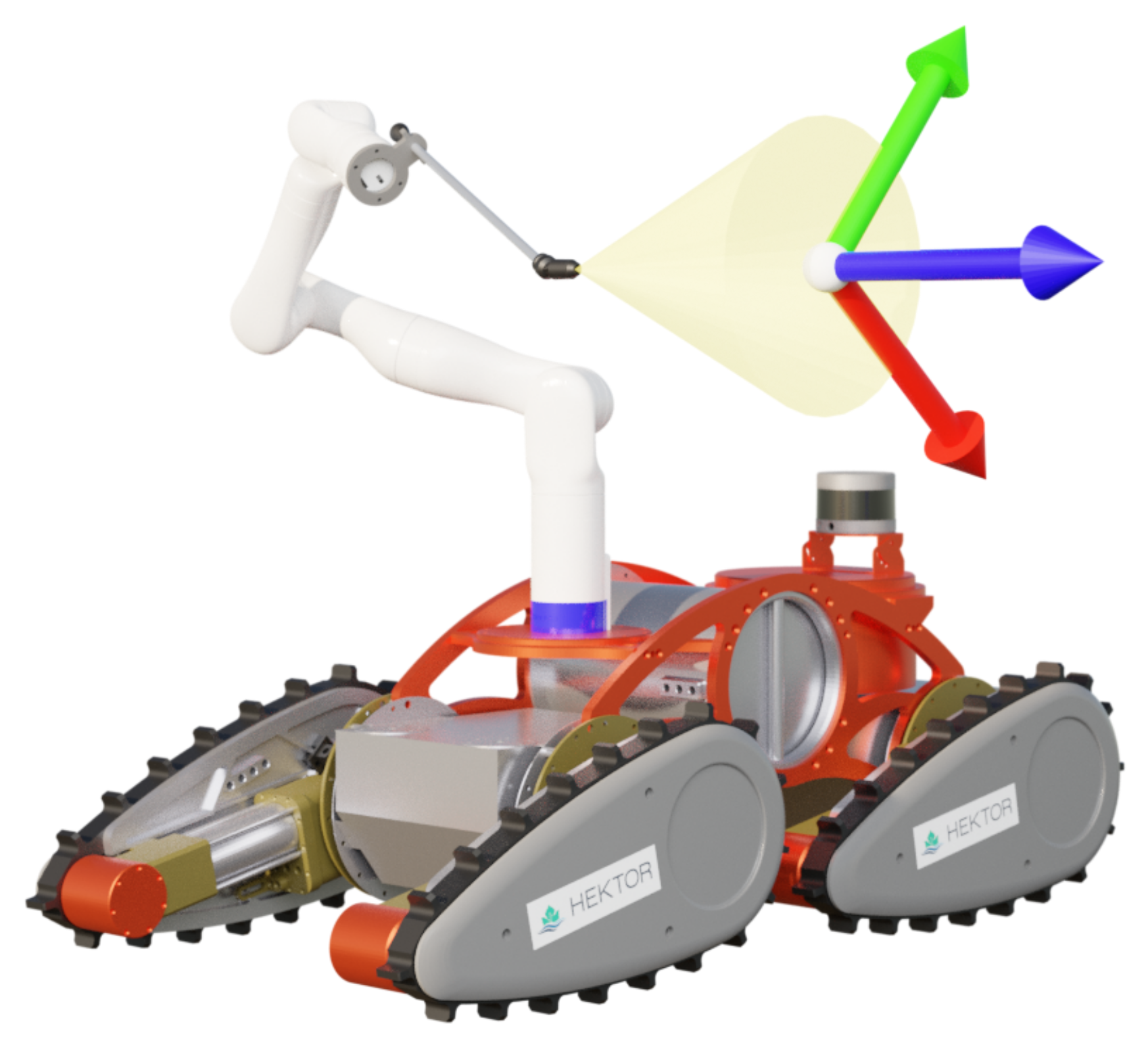}
\caption{
The scenario in this paper involves mounting the spray wand for manual vineyard spraying as the end-effector of a mobile manipulator. The nozzle used to apply the spraying agent is an axis-symmetric tool. 
}
\label{fig:intro_fig}
\end{figure}

Manual agricultural spraying is often performed with a spray wand, a nozzle mounted on the end of a lightweight pole. The nozzle is often mounted at an angle, making it easier for the operator to control both the position and the orientation of the nozzle, and reach high and low areas of the canopy. In the presented work, a spray wand is mounted as the robot arm end-effector (Fig. \ref{fig:intro_fig}), aiming to maintain the advantages of manual spraying while benefiting from increased efficiency and precision of robotic technology.

Our previous work focused on the problem of selecting coordinated control inputs for the vehicle and the robot arm in the same scenario \cite{Vatavuk2022}. The robot arm was controlled in the task space, controlling solely the translational velocity of the spraying frame, depicted in Fig. \ref{fig:intro_fig}, and disregarding its rotation. The reasoning behind this was that, to achieve large enough linear velocities of the spraying frame, and reach high and low areas of the plant, it is not possible to fully control the orientation of the spraying frame. In this paper, a more complete solution to the robot arm control problem is offered, handling the control of the spraying frame as a prioritized 3T2R task.

3T2R tasks, also known as pointing tasks, are frame pose control tasks where all three components of the frame position, and only two components of the frame orientation are considered \cite{Schappler2019}. Since only five degrees of freedom are controlled, a functional redundancy is introduced for robot arms with six degrees of freedom and more. There is extensive research on different approaches to resolving functional redundancies in robot manipulation \cite{From2007, Zlajpah2017, Schappler2019, Zlajpah2021, Zanchettin2011}. 
Tasks performed with axis-symmetric tools, such as robotic welding, paint spraying and drilling, are frequent examples of 3T2R tasks. In robotic drilling for example, both the position and the orientation of the drill bit are important for task execution, but the rotation around the drill bit is not. 

We handle the agricultural spraying task in a similar way, since the rotation around the approach axis of the spraying frame does not effect the application of the spraying agent. However, unlike the drilling task, spraying with a correct spraying frame position and a non ideal approach axis orientation can still be acceptable \cite{From2010, From2011, Zanchettin2011}. 

In \cite{From2011}, From et. al. report that the linear velocity of the paint gun is far more important than its orientation for achieving uniform paint coating. We believe the same to be the case in agricultural spraying, even to a larger extent, since the spraying agent in agricultural spraying is generally less dense than the paint in spray painting applications, and human operators performing the agricultural spraying tasks generally handle the orientation of the nozzle with less care than in the paint spraying applications. 

This insight is handled by introducing a prioritization between translational and rotational components of the 3T2R task. 
Prioritized task space control \cite{deLasa2009, deLasa2010, Wensing2013} replaces commonly used task weighting approach, with hard priorities that are guaranteed to be satisfied. The solution to the lower priority task is found inside the nullspace of a higher priority task. This is performed iteratively, for any number of tasks with different priorities, until a certain task fully constrains the optimization problem. This kind of approach is often referred to as prioritized velocity space inverse kinematics (IK), prioritized instantaneous IK or just prioritized IK \cite{Chiacchio1991, Lillo2019, Moe2016, An2019}. 

In this work, constrained prioritized task space control algorithm presented in \cite{deLasa2010} is used to solve the prioritized velocity space inverse kinematics problem, for the continuous agricultural spraying application. Continuous spraying refers to the task of treating the entire canopy of the plant. Previous robotic approaches to this task mostly used a set of nozzles fixed on the mobile vehicle \cite{Berenstein2010, Berenstein2019, Cantelli2019}, while the robot arm was mostly utilized for selective spraying \cite{Oberti2013, Oberti2016}. Selective spraying refers to the problem of spraying a specific part of the plant, for example a single disease-ridden leaf, or a fruit cluster. A solution to this problem is also presented, handling agricultural selective spraying as a prioritized positional level IK problem. A solver for this purpose based on iterative constrained prioritized task space control is presented. 

\subsection{Contribution}
We present a constrained prioritized 3T2R task control scheme for agricultural spraying, solving the 3T2R control task on both velocity and position levels, prioritizing between its translational and rotational components. Two use cases are discussed in which the velocity and the position level algorithms are applied to continuous and selective agricultural spraying respectively. The implementation of the velocity level prioritized task space control scheme for continuous spraying, and the prioritized positional inverse kinematics solver for selective spraying are discussed in detail. 

\subsection{Paper Organization}
The remainder of this paper is structured as follows: Section II presents the constrained prioritized task space control approach for continuous agricultural spraying. The details of the approach are presented as well as the discussion on the effects of different constrains on the performance. Section III presents the solution to the selective agricultural spraying problem, approached as a prioritized positional inverse kinematics problem. Details of the implementation are presented, and the results and their implications are discussed. Finally, Section IV. concludes the paper with some comments on future work.  
        \section{Constrained Prioritized Task Space Control for Continuous Spraying}
\label{sec:cont_spr}

As already mentioned, continuous spraying refers to the problem of applying the spraying agent to the entire canopy of the plant. Constrained prioritized task space control is used to select joint velocity commands that follow the commanded spraying frame velocity. Velocity of the spraying frame is controlled as a prioritized 3T2R control task, prioritizing its translational over its rotational component.  

\subsection{Velocity Level Prioritized Task Space Control}

\begin{figure*}[!t]
\centering
\includegraphics[width=0.8\textwidth]{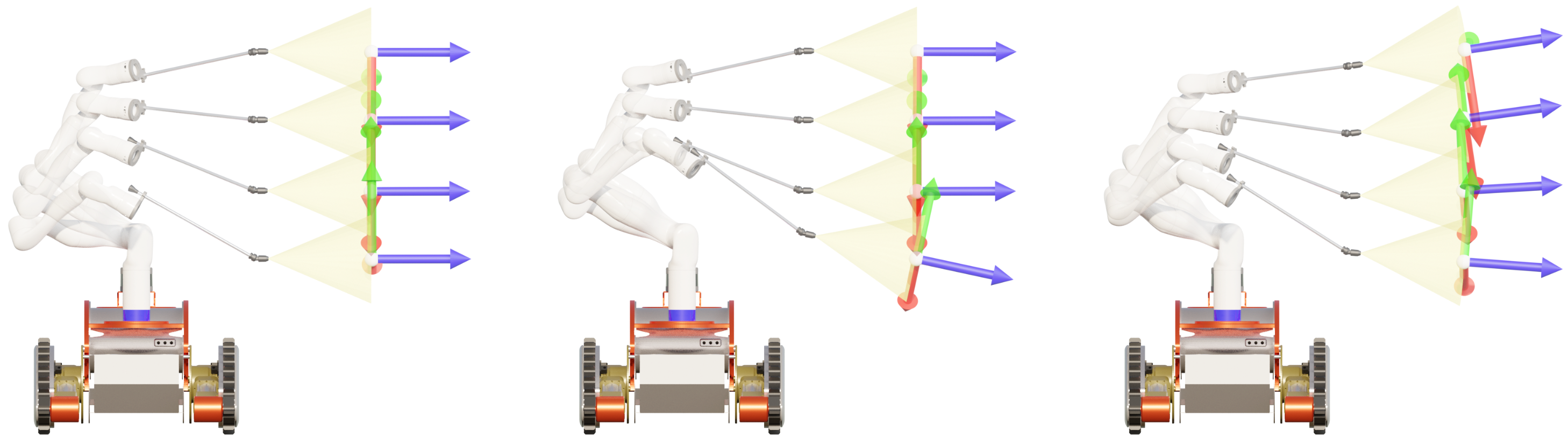}
\caption{ Continuous spraying examples, from left to right: slow spraying, positionally constrained slow spraying, fast spraying. 
}
\label{fig:sideways_fig}
\end{figure*}

Joint velocity commands are selected by solving a constrained prioritized task space control problem \cite{deLasa2010}. The general constrained prioritized task space control problem is defined as:
\begin{equation}
	\begin{aligned}
		h_i = & \ \underset{\boldsymbol{x}}{\text{min}} & & E_i(\boldsymbol{x})\\
		& \ \ \text{s.t.} & & E_k(\boldsymbol{x}) = h_k, \forall k < i\\
		& & & \boldsymbol{A}_{eq}\boldsymbol{x} + \boldsymbol{b}_{eq} = 0\\
		& & & \boldsymbol{A}_{ieq}\boldsymbol{x} + \boldsymbol{A}_{ieq} \geq 0
	\end{aligned}
\end{equation}
where $E_i$ is the quadratic cost function of the $i$-th priority, $h_i$ is the optimal value of that cost function, and $\boldsymbol{A}_{eq}$, $\boldsymbol{b}_{eq}$, $\boldsymbol{A}_{ieq}$ and $\boldsymbol{b}_{ieq}$ are the matrices and vectors describing linear equality and inequality constraints, respectively.

Priorities used for continuous agricultural spraying are:
\begin{itemize}
    \item Translational part of the 3T2R task
    \item Rotational part of the 3T2R task
    \item Desired joint positions
\end{itemize}

The cost function of the first priority has a following form:
\begin{equation} \label{eq:priority_1}
	\begin{aligned}
E_1(\dot{\boldsymbol{q}}) = & \norm{ \boldsymbol{v}_c - \boldsymbol{J}_{T}\dot{\boldsymbol{q}} }^2
	\end{aligned}
\end{equation}
where $\boldsymbol{v_c}$ is the commanded linear velocity of the spraying frame, $\boldsymbol{J}_{T}$ is the translational part of the spraying frame Jacobian, and $\dot{\boldsymbol{q}}$ is the joint velocity vector.
$\boldsymbol{v_c}$ is the output of the MPC solver described in \cite{Vatavuk2022}. Generally, there are multiple joint velocity vectors $\dot{\boldsymbol{q}}$ that result in the commanded linear velocity, and the criterion function of the second priority is selected between those solutions, in the null space of the first priority.

The cost function of the second priority, referring to the rotational part of the 3T2R task, has a following form:
\begin{equation} \label{eq:priority_2}
E_2(\dot{\boldsymbol{q}}) = \norm{ \omega^L_{c,x} - \boldsymbol{J}^L_{R,x}\dot{\boldsymbol{q}} }^2 + \norm{ \omega^L_{c,y} - \boldsymbol{J}^L_{R,y}\dot{\boldsymbol{q}} }^2
\end{equation}
where $\omega^L_{c,x}$ and $\omega^L_{c,y}$ are commanded angular velocities of the spraying frame around its local $x$ and $y$ axes respectively, and $\boldsymbol{J}^L_{R,x}$ and $\boldsymbol{J}^L_{R,y}$ are the corresponding Jacobian matrices. Since the spraying nozzle is an axis-symmetric tool, the angular velocity around its local $z$ axis is not directly controlled.

The final priority, which resolves any redundancy remaining after minimizing the first two priorities, favors such joint velocities $\dot{\boldsymbol{q}}$ that move the arm towards a desired configuration:
\begin{equation} \label{eq:priority_3}
E_3(\dot{\boldsymbol{q}}) = \norm{ \dot{\boldsymbol{q}}_c - \dot{\boldsymbol{q}} }^2
\end{equation}

The commanded joint velocities $\dot{\boldsymbol{q}}_c$ that drive the robot arm towards a desired pose $\boldsymbol{q}_d$ are selected by a proportional controller:
\begin{equation}
\dot{\boldsymbol{q}}_{c} = K_{P,q}(\boldsymbol{q}_d - \boldsymbol{q})
\end{equation}
where $K_{P,q}$ is the controller gain and $\boldsymbol{q}$ is a current joint position vector.

Inequality constraints are used to enforce joint velocity and acceleration limits:
\begin{equation}
\underline{\dot{\boldsymbol{q}}} \leq \dot{\boldsymbol{q}} \leq \overline{\dot{\boldsymbol{q}}}
\end{equation}
\begin{equation}
\label{eq:acc_constr}
\underline{\ddot{\boldsymbol{q}}} \leq \ddot{\boldsymbol{q}} \leq \overline{\ddot{\boldsymbol{q}}}
\end{equation}

Since the prioritized task space control problem deals with joint velocities, equation (\ref{eq:acc_constr}) is replaced with the one in the velocity space:
\begin{equation}
\dot{\boldsymbol{q}}_P + \underline{\ddot{\boldsymbol{q}}}\Delta t \leq \dot{\boldsymbol{q}} \leq \dot{\boldsymbol{q}}_P + \overline{\ddot{\boldsymbol{q}}}\Delta t
\end{equation}
where $\Delta t$ is the control time step, and $\dot{\boldsymbol{q}}_P$ are joint velocities in the previous time step.

\subsection{Commands for the rotational part of the 3T2R task}
Commands for the local angular velocities of the spraying frame $\omega^L_{c,x}$ and $\omega^L_{c,y}$ are calculated using the error between the desired and the current approach axis orientation:
\begin{equation}
err_{\alpha} = \arccos( \boldsymbol{app}_{z} \cdot \boldsymbol{app}_{d,z} )
\end{equation}
\begin{equation}
\boldsymbol{err}_{axis} = \boldsymbol{app}_{z} \times \boldsymbol{app}_{d,z}
\end{equation}
where $\boldsymbol{app}_{z}$ and $\boldsymbol{app}_{z,d}$ are the current and the desired approach axis vectors, respectivelly, $err_{\alpha}$ is the angular distance between the two vectors, and $\boldsymbol{err}_{axis}$ is an axis around which $err_{\alpha}$ acts. 

Angular error vector represented in the local frame is:
\begin{equation}
\boldsymbol{\alpha}^L_{err} = {}_L\boldsymbol{R}^B(err_{\alpha} \cdot \boldsymbol{err}_{axis})
\end{equation}
If the $z$ axis of the frame is considered its approach axis, the $z$ component of $\boldsymbol{\alpha}^L_{err}$ is always zero, and the local angular velocities are calculated as:
\begin{equation}
\boldsymbol{\omega}^L_{c} = K_{P,\omega}\boldsymbol{\alpha}^L_{err} = 
\begin{bmatrix}
    \omega^L_{c,x} \\
    \omega^L_{c,y} \\
    0 \\
\end{bmatrix}
\end{equation}
where $K_{P,\omega}$ is the proportional controller gain.



\subsection{Continuous Spraying Examples}
\begin{figure*}[!ht]
\centering
\includegraphics[width=0.95\textwidth]{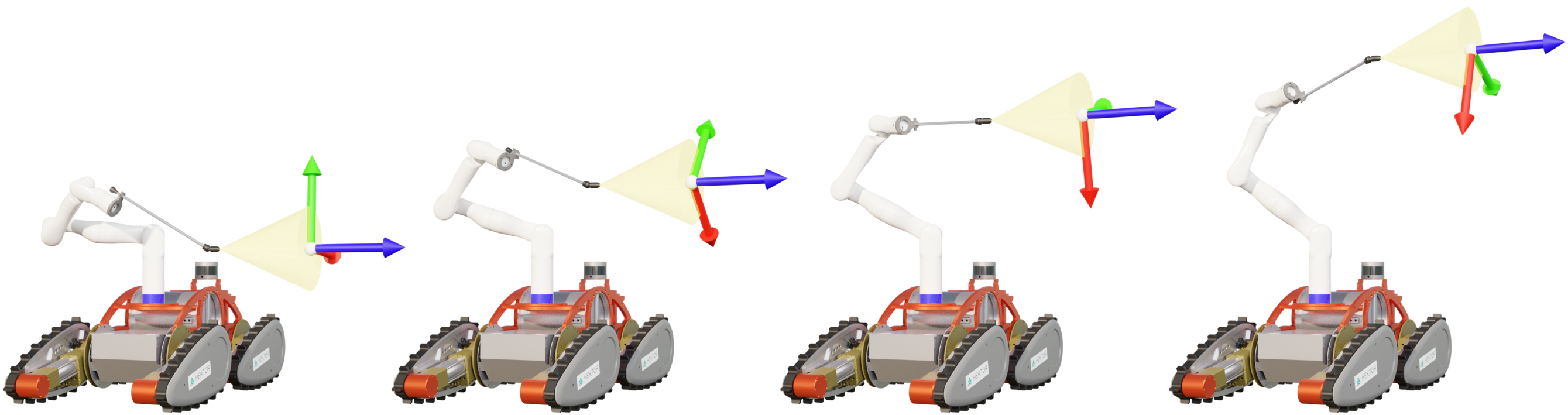}
\caption{
	Spraying frame rotates freely around its approach axis while minimizing the 3T2R task as well as joint movement. 
}
\label{fig:consecutive}
\end{figure*}
The previously described approach was tested on three continuous spraying examples, with different commanded linear velocities and constraints (Fig. \ref{fig:sideways_fig}). In all the examples the spraying frame rotates freely around its approach axis, as a result of 3T2R control (Fig. \ref{fig:consecutive}). Footage of the examples can be seen in the accompanying video\footnote{\url{https://www.youtube.com/watch?v=FRdmGsSCAh4}}. Constrained prioritized task space control solver described in \cite{deLasa2010} by de Lasa et al. was implemented in C++ using OSQP (Operator Splitting Quadratic Program) quadratic programming solver \cite{osqp}. This implementation was used for the experiments, and is available on GitHub\footnote{\url{https://github.com/ivatavuk/ptsc_eigen}}.

First example has a low commanded linear velocity of $0.2$ m/s, resulting in both the linear and rotational velocity being feasible during the entire trajectory. The 3T2R task is followed in its entirety as a result, and the third priority (Eq. (\ref{eq:priority_3})) fully constrains the prioritized optimization problem. 

In the second example, the same commanded linear velocity was used as in the first one, but with an addition of a positional constraint on a nozzle height. The nozzle is not allowed to reach positions lower than $0.3$ m from the robot arm base. During the lower segment of the trajectory, this constraint becomes active, which results in the prioritization between the translational and rotational component of the 3T2R task being noticeable (Fig. \ref{fig:sideways_fig}).    


Third example has a large commanded linear velocity of the spraying frame of $0.8$ m/s, which results in joint velocity and acceleration constraints being reached during the execution of the trajectory. As a consequence, the 3T2R task is not achievable in its entirety, and the third priority is disregarded for the most part of the trajectory. To combat this issue, for the fast trajectory only two priorities are used. The first priority is the same as in the previous examples (Eq. \ref{eq:priority_1}), and the second priority is a weighted combination of the rotational component of the 3T2R task and desired joint movement:

\begin{equation} \label{eq:priority_2_2}
E_2(\dot{\boldsymbol{q}}) = \norm{ \omega^L_{x,d} - \boldsymbol{J}^L_{R,x}\dot{\boldsymbol{q}} }^2 + \norm{ \omega^L_{y,d} - \boldsymbol{J}^L_{R,y}\dot{\boldsymbol{q}} }^2 + w\norm{ \dot{\boldsymbol{q}}_d - \dot{\boldsymbol{q}} }^2
\end{equation}

In this example, the position of the spraying frame follows the commanded linear velocity, while its desired orientation is not achievable due to joint velocity and acceleration constraints.

The utility of the presented method resides in the prioritization between the translational and rotational components of the 3T2R tasks. When the 3T2R task velocity is not feasible in its entirety, due to the constraints posed by the robot arm or due to the custom constraints posed by the user, task priorities are utilized to find an optimal spraying angle.


        \section{Prioritized Positional Inverse Kinematics for Selective Spraying}
Selective agricultural spraying refers to the task of spraying a specific part of the plant, for example a cluster of grapes. This task is handled as a prioritized positional inverse kinematics problem. Prioritization between the translational and rotational components of the 3T2R task used for continuous spraying remains for this use case. 

\subsection{Prioritized Positional Inverse Kinematics Solver}

\begin{figure*}[!ht]
\centering
\includegraphics[width=0.9\textwidth]{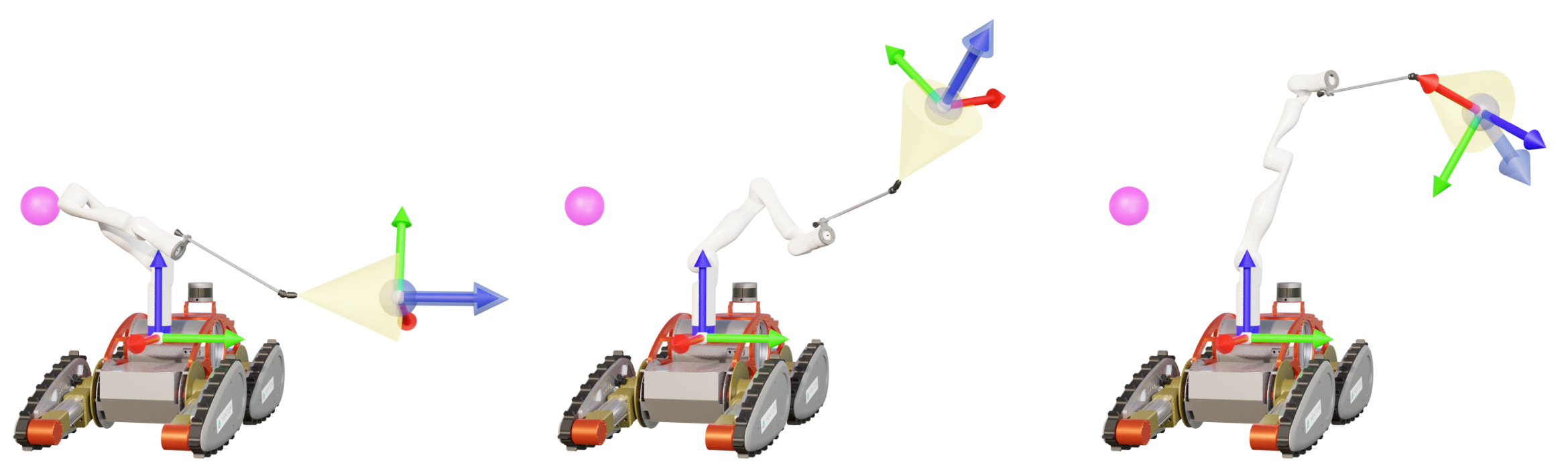}
\caption{Prioritized positional inverse kinematics examples for the task of selective agricultural spraying. Transparent blue sphere and arrow represent the desired position and desired approach axis orientation of the spraying frame respectively, and the transparent purple sphere represents the desired elbow position.}
\label{fig:pik_fig}
\end{figure*}

Prioritized positional inverse kinematics solver implementation is similar to standard numerical inverse kinematics, iteratively solving the velocity level problem. The velocity level problem is solved as a constrained prioritized task-space control problem, as described in section \ref{sec:cont_spr}. While the standard positional inverse kinematics solvers aim to achieve a commanded end-effector pose, the presented solver has the ability of handling multiple, potentially conflicting tasks with different priorities. 

Solver pseudoalgorithm is given in Algorithm \ref{alg:pik}. The algorithm requires an initial guess for joint positions $\boldsymbol{q}_{initial}$. Task errors and Jacobians are calculated based on the current joint positions $\boldsymbol{q}$ and the type of the task. Error gradients are updated for each task as a difference between the task error in current and previous iteration of the algorithm. Task Jacobians and clamped errors are used to construct a prioritized task space control problem. Finally, a solution to the prioritized task space problem is used to update the current joint positions. If the sum of all task error norms or error gradient norms reaches a threshold the problem is considered to be solved. 

\begin{algorithm}
\caption{Positional prioritized inverse kinematics solver.}\label{alg:pik}
$\boldsymbol{q} \gets \boldsymbol{q}_{\text{initial}}$\\
tasks $\gets [ ]$ \\
\While{$\sum \|\boldsymbol{err}_i\| \geq \varepsilon_{e}$ \textbf{and} $\sum \|\nabla \boldsymbol{err}_i\| \geq \varepsilon_{\nabla}$}
{
    \For{$i\gets0$ \KwTo $N$}{
        $\boldsymbol{J}_i \gets \text{getTaskJacobian}(\boldsymbol{q}, \text{tasktype}_i)$\\
        $\boldsymbol{err}_i \gets \text{getTaskError}(\boldsymbol{q}, \text{tasktype}_i)$\\
        $\nabla \boldsymbol{err}_i \gets \text{updateGradient}(\boldsymbol{err}_i)$\\
        $\boldsymbol{err}_i \gets \text{clampTaskError}(\boldsymbol{err}_i, \text{tasktype}_i)$\\
        tasks.\text{insert}$(\boldsymbol{J}_i, \boldsymbol{err}_i)$\\
    }
    $\boldsymbol{q} \gets \text{solvePTSC}(\text{tasks}, \text{constraints})$\\
    tasks.\text{clear}()\\
}
\end{algorithm}
 
Prioritized inverse kinematics library for ROS is available on GitHub\footnote{\url{https://github.com/ivatavuk/pik_ros}}. MoveIt is used to calculate the Jacobians of the specified frames, which must be present in the URDF file of the MoveIt planning group. 

The Jacobians obtained with MoveIt are modified to support any of the following tasks: 
\begin{itemize}
    \item Frame pose task
    \item Frame position task
    \item Frame orientation task
    \item Frame approach axis vector task
\end{itemize}
These task types correspond to the $tasktype_i$ variable in the pseudoalgorithm \ref{alg:pik}. A frame pose task Jacobian is the standard Jacobian matrix, and frame position and orientation task Jacobians correspond to the first and last three rows of the frame pose Jacobian. The Jacobian and the error for the frame approach axis vector task are calculated as described in section \ref{sec:cont_spr}. 

\subsection{Selective Spraying Examples}
Tasks used in selective agricultural spraying examples are, with decreasing priorities:
\begin{itemize}
    \item Spraying frame position task
    \item Spraying frame approach axis orientation task
    \item Elbow frame position task
\end{itemize}

Like for the continuous spraying, there is a prioritization of the spraying frame position over its approach axis orientation, which correspond to the translational and rotational components of the 3T2R task. The third priority, which fully constrains the positional inverse kinematics problem is the desired elbow frame position.

The solver was tested on three different examples seen in Fig. \ref{fig:pik_fig}, with desired values for all the tasks given in table \ref{tab:pik_examples}. Tasks are set up in such a way that in the first two examples the desired values of the full 3T2R task are feasible, and the elbow position task fully constrains the problem, and in the last example only the position of the spraying frame is feasible (Fig. \ref{fig:pik_fig}). 

\begin{table}[]
    \centering
    \begin{tabular}{c | c c c c}
         Example & \makecell{ Spraying frame \\ position [m] }  & \makecell{ Spraying frame \\ approach axis \\ vector } & \makecell{ Elbow \\ position [m]} \\
         \hline
         1 & [0.4 1.0 0.2] & [0 1 0] & [0.0 -0.5 0.5] \\
         2 & [0.4 1.0 0.8] & [0.511 0.511 0.69] & [0.0 -0.5 0.5] \\
         3 & [0.4 1.0 0.8] & [0.577 0.577 -0.577] & [0.0 -0.5 0.5]
    \end{tabular}
    \caption{Desired values for prioritized tasks used in the examples.}
    \label{tab:pik_examples}
\end{table}

The description of solver performance for the examples is given in Tab. \ref{tab:pik_results}. All experiments were conducted on a $2.2$\textit{GHz} Intel Core i7 processor. It can be noticed that the third example takes the largest amount of time to be solved, which is due to the solution being close to the robot arm singularity. 

\begin{table}[]
    \centering
    \begin{tabular}{c | c c c c}
         Example & \makecell{ Task 1 \\ err [m] }  & \makecell{ Task 2 \\ err [rad]} & \makecell{ Task 3 \\ err [m]} & \makecell{ Time \\ \textnormal{[ms]} } \\
         \hline
         1 & 0.00020 & 0.00019 & 0.3525 & 11.09 \\
         2 & 0.00054 & 0.00067 & 0.7222 & 19.86 \\
         3 & 0.00247 & 0.2052 & 0.6253 & 30.07 
    \end{tabular}
    \caption{Task errors and calculation time for the examples.}
    \label{tab:pik_results}
\end{table}

Parameters for the solver used in the examples are:
\begin{itemize}
    \item Use constrained optimization = $True$
    \item Error norm gradient threshold = $1\times 10^{-3}$
    \item Change in joint angle constraint = $10$ [$^{\circ}$]
    \item Use solution polishing = $True$
    \item Polish error norm gradient threshold = $1\times 10^{-2}$
    \item Polish change in joint angle constraint = $3$ [$^{\circ}$]
    \item Positional clamp magnitude = $0.3$ [m]
    \item Orientational clamp magnitude = $30$  [$^{\circ}$]
\end{itemize}

Solution polishing refers to the usage of smaller change in joint angle constraint when the solver is close to the solution, which is detected as \textit{polish error norm gradient threshold} being reached.  

All three tasks are not feasible in any of the given examples, so the solver considers the positional prioritized IK problem solved once task error gradients reach a specified threshold. For most prioritized inverse kinematics applications the same would be the case, as the main strength of this approach is its ability to handle conflicting, infeasible tasks with clearly defined priorities. 

	\section{Conclusion and Future Work}
For the task of robotic agricultural spraying, and for robotic spraying in general, the position of the spraying frame is more important than its orientation to ensure satisfactory spray coverage. We propose a solution where constrained prioritized optimization is used for velocity and positional level 3T2R task control, which corresponds to continuous and selective agricultural spraying tasks, respectively. Prioritized task space control and prioritized positional inverse kinematics are described in detail. Positional inverse kinematics are solved using iterative constrained prioritized task space control. In the future work, the prioritized IK framework is planned to be expanded to allow for more task types, such as preferred joint positions, manipulability maximization task and others. Voxel based obstacle avoidance is also planned to be included. We plan to explore the applicability of the framework for different robot control tasks. Prioritized positional inverse kinematics could have a number of applications, most interesting ones including high dimensional floating base robotic systems, which could have a high number of prioritized conflicting tasks. The utility of the prioritized optimization described in this paper for trajectory planning also remains to be explored.

	\subsubsection*{Acknowledgments}
	Research work presented in this article has been supported by the project Heterogeneous autonomous robotic system in viticulture and mariculture (HEKTOR) financed by the European Union through the European Regional Development Fund-The Competitiveness and Cohesion Operational Programme (KK.01.1.1.04.0036).
	\nocite{*}
	\bibliographystyle{ieeetr}
	\bibliography{bibliography/asdf}
	
\end{document}